\documentclass[11pt]{article}
\usepackage{eacl2017}
\usepackage{times}
\usepackage{url}
\usepackage{latexsym}

\eaclfinalcopy

\usepackage[table]{xcolor}

\usepackage{tikz}
\usepackage{pifont}
\usepackage[inline]{enumitem}
\usetikzlibrary{positioning}
\usetikzlibrary{calc}
\usetikzlibrary{fit}
\usetikzlibrary{decorations.text}
\usetikzlibrary{shapes.misc}
\usetikzlibrary{shapes.geometric,arrows}
\usepackage{colortbl}

\newcommand{\ie}{i.e.\xspace}

\newcommand{\system}[1]{\texttt{#1}\xspace}

\newcommand{\equref}[1]{Equation~(\ref{#1})\xspace}

\newcommand{\bb}[1]{\mathbb{#1}}
\newcommand{\R}{\bb{R}}
\newcommand{\mat}[2][]{\boldsymbol{#2}_{#1}}

\renewcommand{\vec}[2][]{\boldsymbol{#2}^{#1}}

\newcommand{\softmax}{\textrm{softmax}}

\newcommand{\T}{\mathstrut\scriptscriptstyle\top}

\newcommand{\memnn}{\system{MemN2N}}

\newcommand{\RNum}[1]{\uppercase\expandafter{\romannumeral #1\relax}}

\title{Dialog state tracking, \\a machine reading approach using Memory Network }

\author{Julien Perez {} {}\\
  Xerox Research Centre Europe \\
  Grenoble, France \\
  {\tt {julien.perez@xrce.xerox.com}} {} {} {} {} {} {} {} \\\And
{}Fei Liu \thanks{{} {} Work carried out as an intern at XRCE}{} {} \\
 {} The University of Melbourne \\
 {} Victoria, Australia \\
  {} {} {} {} {} {}{\tt {fliu3@student.unimelb.edu.au} {} {}} \\}

\usepackage{mathptmx}       
\usepackage{helvet}         
\usepackage{courier}        
\usepackage{type1cm}        
\usepackage{makeidx}         
\usepackage{graphicx}        
\usepackage[bottom]{footmisc}

\usepackage{booktabs}
\usepackage{xspace}

\usepackage{amsmath, amsfonts}

\DeclareMathOperator*{\argmax}{arg\,max}

\begin{document} 
\maketitle
 
\begin{abstract}
In an end-to-end dialog system, the aim of dialog state tracking is to accurately estimate a compact representation of the current dialog status from a sequence of noisy observations produced by the speech recognition and the natural language understanding modules. This paper introduces a novel method of dialog state tracking based on the general paradigm of machine reading and proposes to solve it using an End-to-End Memory Network, \memnn, a memory-enhanced neural network architecture. We evaluate the proposed approach on the second Dialog State Tracking Challenge (DSTC-2) dataset. The corpus has been converted for the occasion in order to frame the hidden state variable inference as a question-answering task based on a sequence of utterances extracted from a dialog. We show that the proposed tracker gives encouraging results. Then, we propose to extend the DSTC-2 dataset and the definition of this dialog state task with specific reasoning capabilities like counting, list maintenance, yes-no question answering and indefinite knowledge management. Finally, we present encouraging results using our proposed \memnn based tracking model.
\end{abstract} 

\section{Introduction}

One of the core components of state-of-the-art and industrially deployed dialog systems is a dialog state tracker. Its purpose is to provide a compact representation of a dialog produced from past user inputs and system outputs which is called the dialog state. The dialog state summarizes the information needed to successfully maintain and finish a dialog, such as users' goals or requests. In the simplest case of a so-called {\it slot-filling schema}, the state is composed of a predefined set of variables with a predefined domain of expression for each of them. As a matter of fact, in the recent context of end-to-end trainable machine learnt dialog systems, state tracking remains a central element of such architectures \cite{WenGMRSUVY16}. Current models, mainly based on the principle of discriminative learning, tend to share three common limitations. First, the tracking task is perform using a fixed window of the past dialog utterances as support for decision. Second, the possible correlations between the set of tracked variables are not leveraged and individual trackers tend to be learnt independently. Third, the tracking task is summarized as the capability of infering values for a predefined set of latent variables. Starting from these observations, we propose to formalize the task of state tracking as a particular instance of machine reading problem. Indeed, these formalization and the proposed resolution model called \memnn \cite{WestonBCM15} allow to define a tracker that is be able to decide at the utterance level on the basis on the current entire dialog. Indeed, the model learns to focus its attention on the meaningful parts of the dialog regarding the currently asked slot and can eventually capture possible correlation between slots. As far as our knowledge goes, it is the first attempt to explicitly frame the task of dialog state tracking as a machine reading problem. Finally, such formalization allows for the implementation of approximate reasoning capability that has been shown to be crucial for any machine reading tasks \cite{WestonBCM15} while extending the task from slot instantiation to question answering. This paper is structured as follows, Section \ref{sec:pb} recalls the main definitions associated to transactional dialogs and describes the associated problem of statistical dialog state tracking with both the generative and discriminative approaches. At the end of this section, the limitations of the current models in terms of necessary annotations and reasoning capabilities are addressed. Then, Section \ref{sec:prop} depicts the proposed machine reading model for dialog state tracking and proposes to extend a state of the art dialog state tracking dataset, {\it DSTC-2}, to several simple reasoning capabilities. Section \ref{sec:xp} illustrates the approach with experimental results obtained using a state of the art benchmark for dialog state tracking.
\section{Dialog state tracking}
\label{sec:pb}

\subsection{Main Definitions}
A dialog state tracking task is formalized as follows: at each turn of a dyadic dialog, the dialog agent chooses a dialog act $d$ to express and the user answers with an utterance $u$. In the simplest case, the dialog state at each turn is defined as a distribution over a set of predefined variables, which define the structure of the state \cite{Williams05}. This classic state structure is commonly called {\it slot filling} or {\it semantic frame}. In this context, the state tracking task consists of estimating the value of a set of predefined variables in order to perform a procedure or transaction which is the purpose of the dialog. Typically, a natural language understanding module processes the user utterance and generates an N-best list $o = \{ (d_1, f_1), \ldots , (d_n, f_n)\}$, where $d_i$ is the hypothesized user dialog act and $f_i$ is its confidence score. Various approaches have been proposed to define dialog state trackers. The traditional methods used in most commercial implementations use hand-crafted rules that typically rely on the most likely result from an NLU module \cite{YehDJRRPLTBM14} and hardly models uncertainty. However, these rule-based systems are prone to frequent errors as the most likely result is not always the correct one ~\cite{Williams14}. 

More recent methods employ statistical approaches to estimate the posterior distribution over the dialog states allowing them to leverage the uncertainty of the results of the NLU module. In the simplest case where no ASR and NLU modules are employed, as in a text based dialog system \cite{Henderson13a}, the utterance is taken as the observation using a so-called bag of words representation. If an NLU module is available, standardized dialog act schemas can be considered as observations \cite{bunt10}. Furthermore, if prosodic information is available from the ASR component of the dialog system \cite{MiloneR03}, it can also be considered as part of the observation definition. A statistical dialog state tracker maintains, at each discrete time step $t$, the probability distribution over states, $b(s_t)$, which is the system's {\it belief} over the state. The actual slot filling process is composed of the cyclic tasks of {\it information gathering} and integration, in other words -- {\it dialog state tracking}. In such framework, the purpose is to estimate as early as possible in the course of a given dialog the correct instantiation of each variable. In the following, we will assume the state is represented as a set of variables with a set of known possible values associated to each of them. Furthermore, in the context of this paper, only the bag of words has been considered as an observation at a given turn but dialog acts or detected named entity provided by an SLU module could have also been incorporated.

Two statistical approaches have been considered for maintaining the distribution over a state given sequential NLU output. First, the discriminative approach aims to model the posterior probability distribution of the state at time $t+1$ with regard to state at time $t$ and observations $z_{1:t}$. Second, the generative approach attempts to model the transition probability and the observation probability in order to exploit possible interdependencies between hidden variables that comprise the dialog state.

\subsection{Generative Dialog State Tracking}

A generative approach to dialog state tracking computes the belief over the state using Bayes' rule, using the belief from the last turn $b(s_{t-1})$ as a prior and the likelihood given the user utterance hypotheses $p(z_t|s_t)$, with $z_t$ the observation gathered at time $t$. In prior works \cite{Williams05}, the likelihood is factored and some independence assumptions are made: $b_t \propto \sum_{s_{t-1},z_t} p(s_t|z_t, s_{t-1}) p(z_t|s_{t-1})  b(s_{t-1})$. A typical generative model uses a factorial hidden Markov model \cite{gj97}. In this family of approaches, scalability is considered as one of the main issues. One way to reduce the amount of computation is to group the states into partitions, as proposed in the Hidden Information State (HIS) model \cite{GasicY11}. Other approaches to cope with the scalability problem in dialog state tracking is to adopt a factored dynamic Bayesian network by making conditional independence assumptions among dialog state components, and then using approximate inference algorithms such as loopy belief propagation \cite{ThomsonY10} or a blocked Gibbs sampling as \cite{RauxM11}. To cope with such limitations, discriminative methods of state tracking presented in the next part of this section aim at directly model the posterior distribution of the tracked state using a choosen parametric form.

\subsection{Discriminative Dialog State Tracking}
\label{sec:paramsoa}

The discriminative approach of dialog state tracking computes the belief over a state via a parametric model that directly represents the belief $b(s_{t+1}) = p(s_{s+1} | s_t, z_t)$. For example, Maximum Entropy has been widely used in the discriminative approach \cite{MetallinouBW13}. It formulates the belief as follows: $b(s) = P(s|x) = \eta.e^{w^T\phi(x,s)} $, where $\eta$ is the normalizing constant, $x = (d^u_1, d^m_1, s_1, \dots, d^u_t, d^m_t, s_t)$ is the history of user dialog acts, $d^u_i, i \in \{1,\ldots,t\}$, the system dialog acts, $d^m_i, i \in \{1,\ldots,t\}$, and the sequence of states leading to the current dialog turn at time $t$. Then, $\phi(.)$ is a vector of feature functions on $x$ and $s$. Finally, $w$ is the set of model parameters to be learned from annotated dialog data. Finally, deep neural models, performing on a sliding window of features extracted from previous user turns, have also been proposed in \cite{Henderson14,SWTY16}. Of the current litterature, this family of approaches have proven to be the most efficient for publicly available state tracking datasets. Recently, deep learning based models implementing this strategy \cite{SWTY16,Henderson2014d,WilliamsRH16a} have shown state of the art results. This approaches tends to leverage unsupervised training word representation \cite{MikolovSCCD13}.

\subsection{Current Limitations}

\begin{figure}[tb]
\centering
\resizebox{.5\textwidth}{!}{
\includegraphics{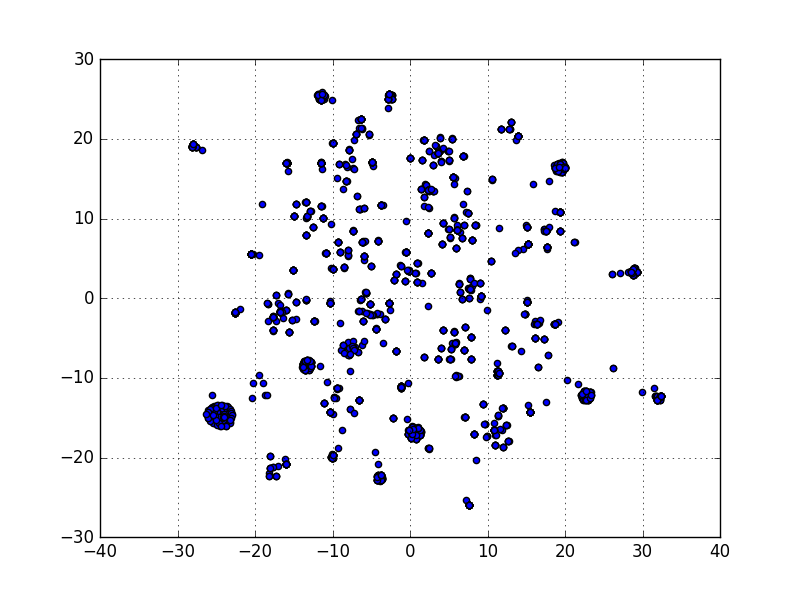}
}
\caption{\label{fig:tsne}T-SNE transformation of the final state of DSTC-2 train set.}
\end{figure}

Using error analysis \cite{HendersonTW14}, three limitations can be observed in the application of these inference approaches. First, current models tend to fail at considering long-tail dependencies that occurs on dialogs. For example, coreferences, inter-utterances informations and correlations between slots have been shown to be difficult to handle even with the usage of recurrent network models \cite{Henderson2014d}. To illustrate the inter-slot correlation, Figure \ref{fig:tsne} depicted the t-SNE \cite{Maaten2008} projected final state of the dialog of the DSTC-2 training set. On the other hand, reasoning capabilities, as required in machine reading applications \cite{PoonD10,EtzioniBC07,BerantSCLHHCM14,WestonBCM15} remain absent in these classic formalizations of dialog state tracking. Finally, tracking definition is limited to the capability to instantiate a predefined set of slots. In the next section, we present a model of dialog state tracking that aims at leveraging the current advances of \memnn, a memory-enhanced neural networks and their approximate reasoning capabilities that seems particularly adapted to the sequential, long range dependency equipped and sparse nature of complex dialog state tracking tasks. Furthermore, this model allows to relax the hypothesis of strict utterance-level annotation that does not corresponds to common pratices in industrial applications of transactional conversational user interfaces where annotations tend to be placed at a multi-utterance level or full-dialog level only. 

\section{Machine Reading Formulation of Dialog State Tracking}
\label{sec:prop}

We propose to formalize the dialog state tracking task as a machine reading problem \cite{EtzioniBC07,BerantSCLHHCM14}. In this section, we recall the main definitions of the task of machine reading, then describes the \memnn, a memory-enhanced neural network architectures proposed to handle such tasks in the context of dialogs. Finally, we formalize the task of dialog state tracking as a machine reading problem and propose to solve it using a memory-enhanced neural architecture of inference.

\subsection{Machine Reading}

The task of textual understanding has recently been formulated as a supervised learning problem \cite{KumarISBEPOGS15,HermannKGEKSB15}. This task consists in estimating the conditional probability $p(a|d, q)$ of an answer $a$ to a question $q$ where $d$ is a document. Such an approach requires a large training corpus of \{Document - Query - Answer\} triples and until now such corpora have been limited to hundreds of examples \cite{RichardsonBR13}.  In the context of dialog state tracking, it can be understood as the capability of inferring a set of latent values $l$ associated with a set of variables $v$ related to a given dyadic or multi-party conversation $d$, from direct correlation and/or reasoning, using the course of exchanges of utterances, $p(l | d , v)$.

State updates at an utterance-level are rarely provided off-the-shelf from a production environment. In these environments, annotation is often performed afterhand for the purpose of logging, monitoring or quality assessment.  In the limit cases, as in human-to-human dialog systems, dialog-level annotations remains a common pratices of annotation especially in personal assistance, customer care dialogs and, in a more general sense, industrial application of transactional conversational user interfaces. Another frequent setting consist of informing the state after a given number of utterance exchange between the locutors. So an additional effort of specific annotation is often needed in order to train a state of the art statistical state tracking model \cite{HendersonTW14}. In that sense, formalizing dialog state tracking at a subdialog level in order to infer hidden state variables with respect to a list of utterances started from the first one to any given utterance of a given dialog seems particularly appropriate. In the context of dialog state tracking challenges, the {\it DSTC-4} dialog corpus have been designed in such purpose but only consists of 22 dialogs. Concerning the {\it DSTC-2} corpus, the training data contains 2207 dialogues (15611 turns) and the test set consists of 1117 dialogues \cite{WilliamsRH16a}. This dataset is more suitable for our experiments.

For these reasons, the machine reading paradigm becomes a promising formulation for the general problem of dialog state tracking. Furthermore, current approaches and available datasets for state tracking do not explicitly cover reasoning capabilities such as temporal and spatial reasoning, couting, sorting and deduction. We suggest that in the future dataset dialogs expressing such specific abilities should be developed. In this last part, several reasoning enhancements are suggested to the {\it DSTC-2} dataset.

\subsection{End-to-End Memory Networks}

The \memnn architecture, introduced by~\cite{WestonBCM15}, consists of two main components: supporting memories and final answer prediction. Supporting memories are in turn comprised of a set of input and output memory representations with memory cells. The input and output memory cells, denoted by $\vec{m}_{i}$ and $\vec{c}_{i}$, are obtained by transforming the input context $x_{1},\ldots,x_{n}$ (i.e a set of utterances) using two embedding matrices $\mat{A}$ and $\mat{C}$ (both of size $d \times |V|$ where $d$ is the embedding size and $|V|$ the vocabulary size) such that $\vec{m}_{i} = \mat{A}\Phi(x_{i})$ and $\vec{c}_{i} = \mat{C}\Phi(x_{i})$ where $\Phi(\cdot)$ is a function that maps the input into a bag of dimension $|V|$. 

Similarly, the question $q$ is encoded using another embedding matrix $\mat{B} \in \R^{d \times |V|}$, resulting in a question embedding $\vec{u} = \mat{B}\Phi(q)$. The input memories $\{\vec{m}_{i}\}$, together with the embedding of the question $\vec{u}$, are utilized to determine the relevance of each of the stories in the context, yielding in a vector of attention weights 
\begin{equation}
p_{i} = \softmax(\vec[\T]{u}\vec{m}_{i})
\end{equation}
where $\softmax(a_{i}) = \dfrac{e^{a_{i}}}{\sum_{i}e^{a_{i}}}$. Subsequently, the response $\vec{o}$ from the output memory is constructed by the weighted sum: 
\begin{equation}
\vec{o} = \sum_{i}p_{i}\vec{c}_{i}
\end{equation}
Other models of parametric encoding for the question and the document have been proposed in \cite{KumarISBEPOGS15}. For the purpose of this presentation, we will keep with definition of $\Phi$. 

For more difficult tasks requiring multiple supporting memories, the model can be extended to include more than one set of input/output memories by stacking a number of memory layers. In this setting, each memory layer is named a hop and the $(k+1)^{\textrm{th}}$ hop takes as input the output of the $k^{\textrm{th}}$ hop:
\begin{equation}
\label{equ:memnn}
\vec[k+1]{u} = \vec[k]{o} + \vec[k]{u}
\end{equation}

Lastly, the final step, the prediction of the answer to the question $q$, is performed by 
\begin{equation}
\label{equ:finalpred}
\hat{\vec{a}} = \softmax(\mat{W}(\vec[K]{o} + \vec[K]{u}))
\end{equation}
where $\hat{\vec{a}}$ is the predicted answer distribution, $\mat{W} \in \R^{|V| \times d}$ is a parameter matrix for the model to learn and $K$ the total number of hops.

Two weight tying schemes of the embedding matrices have been introduced in \cite{WestonBCM15}:
\begin{enumerate}[noitemsep,topsep=0pt]
	\itemsep0em
	\item \textbf{Adjacent:} the output embedding matrix in the $k^{\textrm{th}}$ hop is shared with the input embedding matrix in the $(k+1)^{\textrm{th}}$ hop, \ie, $\mat{A}^{k+1} = \mat{C}^{k}$ for $k \in \{1, K-1\}$. Also, the weight matrix $\mat{W}$ in \equref{equ:finalpred} is shared with the output embedding matrix in the last memory hop such that $\mat{W}^{\T} = \mat{C}^{K}$.
	\item \textbf{Layer-wise:} all the weight matrices $\mat{A}^{k}$ and $\mat{C}^{k}$ are shared across different hops, \ie, $\mat{A}^{1} = \mat{A}^{2} = \ldots = \mat{A}^{K}$ and $\mat{C}^{1} = \mat{C}^{2} = \ldots = \mat{C}^{K}$.
\end{enumerate}

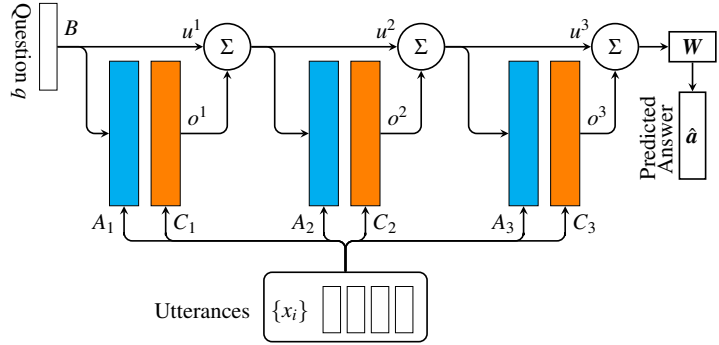
\begin{figure*}[tb]
\begin{center}
\resizebox{.6\textwidth}{!}{
\begin{tikzpicture}

\node[draw,thick,align=center,minimum height=1.2cm,minimum width=2.8cm,rounded corners] at (5.8,1.0) (stories) {};
\node[anchor=north west,shift={(0mm,2.5mm)}] (stories_xi) at (stories.west){$\{x_i\}$};
\node[draw,align=center,minimum height=0.8cm,minimum width=0.3cm] [right=0.1of stories_xi,shift={(0mm,0.3mm)}] (stories_sent1) {};
\node[draw,align=center,minimum height=0.8cm,minimum width=0.3cm] [right=0.1of stories_sent1] (stories_sent2) {};
\node[draw,align=center,minimum height=0.8cm,minimum width=0.3cm] [right=0.1of stories_sent2] (stories_sent3) {};
\node[draw,align=center,minimum height=0.8cm,minimum width=0.3cm] [right=0.1of stories_sent3] (stories_sent4) {};
\node[anchor=north west,shift={(-20mm,2mm)}] (sentences_label) at (stories.west){Utterances};

\node[draw,align=center,minimum height=1.5cm,minimum width=0.3cm] at (0.7,5.5) (question) {};
\node[anchor=north west,shift={(0mm,9mm)},rotate=-90] (question_label) at (question.west){Question $q$};

\node[draw,align=center,minimum height=2.5cm,minimum width=0.5cm,fill=cyan] at (2.0,4.0) (A1) {};
\node[draw,align=center,minimum height=2.5cm,minimum width=0.5cm,fill=orange] [right=0.2of A1] (C1) {};

\node[draw,thick,align=center,circle,minimum size=0.8cm] [right=2.5 of question] (Sum1) {$\Sigma$};

\draw [rounded corners,thick,->,>=stealth] (question) -- ($(question.east) + (0.5,0.0)$) node [midway,above] (B_label) {$\mat{B}$} |- (A1);
\draw [rounded corners,thick,->,>=stealth] (stories) |- ($(A1.south) + (0.0,-0.6)$) -| (A1.south) node [left,shift={(0mm,-3mm)}] (A1_label) {$\mat[1]{A}$};
\draw [rounded corners,thick,->,>=stealth] (stories) |- ($(C1.south) + (0.0,-0.6)$) -| (C1.south) node [right,shift={(0mm,-3mm)}] (C1_label) {$\mat[1]{C}$};

\draw [rounded corners,thick,->,>=stealth] (question) -- (Sum1) node [above,shift={(-6mm,0mm)}] (u1_label) {$\vec[1]{u}$};


\draw [rounded corners,thick,->,>=stealth] (C1.east) -- ($(C1.east) + (0.3,0.0)$)  node [above,shift={(0mm,0mm)}] () {$\vec[1]{o}$} -| (Sum1.south);
%

\node[draw,align=center,minimum height=2.5cm,minimum width=0.5cm,fill=cyan] [right=2.2 of C1] (A2) {};
\node[draw,align=center,minimum height=2.5cm,minimum width=0.5cm,fill=orange] [right=0.2of A2] (C2) {};

\node[draw,thick,align=center,circle,minimum size=0.8cm] [right=2.5 of Sum1] (Sum2) {$\Sigma$};

\draw [rounded corners,thick,->,>=stealth] (Sum1) -- ($(Sum1.east) + (0.4,0.0)$) |- (A2);
\draw [rounded corners,thick,->,>=stealth] (stories) |- ($(A2.south) + (0.0,-0.6)$) -| (A2.south) node [left,shift={(0mm,-3mm)}] (A2_label) {$\mat[2]{A}$};
\draw [rounded corners,thick,->,>=stealth] (stories) |- ($(C2.south) + (0.0,-0.6)$) -| (C2.south) node [right,shift={(0mm,-3mm)}] (C2_label) {$\mat[2]{C}$};

\draw [rounded corners,thick,->,>=stealth] (Sum1) -- (Sum2) node [above,shift={(-6mm,0mm)}] (u2_label) {$\vec[2]{u}$};
%
%
\draw [rounded corners,thick,->,>=stealth] (C2.east) -- ($(C2.east) + (0.3,0.0)$)  node [above,shift={(0mm,0mm)}] () {$\vec[2]{o}$} -| (Sum2.south);
%

\node[draw,align=center,minimum height=2.5cm,minimum width=0.5cm,fill=cyan] [right=2.2 of C2] (A3) {};
\node[draw,align=center,minimum height=2.5cm,minimum width=0.5cm,fill=orange] [right=0.2of A3] (C3) {};

\node[draw,thick,align=center,circle,minimum size=0.8cm] [right=2.5 of Sum2] (Sum3) {$\Sigma$};

\draw [rounded corners,thick,->,>=stealth] (Sum2) -- ($(Sum2.east) + (0.4,0.0)$) |- (A3);
\draw [rounded corners,thick,->,>=stealth] (stories) |- ($(A3.south) + (0.0,-0.6)$) -| (A3.south) node [left,shift={(0mm,-3mm)}] (A3_label) {$\mat[3]{A}$};
\draw [rounded corners,thick,->,>=stealth] (stories) |- ($(C3.south) + (0.0,-0.6)$) -| (C3.south) node [right,shift={(0mm,-3mm)}] (C3_label) {$\mat[3]{C}$};

\draw [rounded corners,thick,->,>=stealth] (Sum2) -- (Sum3) node [above,shift={(-6mm,0mm)}] (u3_label) {$\vec[3]{u}$};

\draw [rounded corners,thick,->,>=stealth] (C3.east) -- ($(C3.east) + (0.3,0.0)$)  node [above,shift={(0mm,0mm)}] () {$\vec[3]{o}$} -| (Sum3.south);


\node[draw,thick,align=center,minimum width=0.8cm] [right=0.5 of Sum3] (W) {$\textbf{\textit{W}}$};

\node[draw,thick,align=center,minimum height=1.5cm] [below=0.5 of W] (a) {$\hat{\textbf{\textit{a}}}$};
\node[anchor=north west,shift={(-8mm,-9mm)},rotate=90] (answer_label) at (a.west){Predicted};
\node[anchor=north west,shift={(-5mm,-8mm)},rotate=90] (answer_label) at (a.west){Answer};

\draw [rounded corners,thick,->,>=stealth] (Sum3.east) -- (W.west);
\draw [rounded corners,thick,->,>=stealth] (W.south) -- (a.north);

\end{tikzpicture}
}
\end{center}
\caption{\label{fig:resmemn2n}Illustration of the proposed \memnn based state dialog tracker model with $3$ hops.}
\end{figure*}

In the next section, we show how the task of dialog state tracking can be formalized as machine reading task and solved using such memory enhanced model.

\subsection{Dialog Reading Model for State Tracking}
\label{sec:model}

In this section, we formalize dialog state tracking using the paradigm of machine reading. As far as our knowledge goes, it is the first attempt to apply this approach and develop a specific dataset format, detailed in Section \ref{sec:xp}, from an existing and publicly available dialog state tracking challenge dataset to fulfill this task. Assuming (1) a dyadic dialog $d$ composed of a list of utterances, (2) a state composed with (2a) a set of variables $v_i$ with $i=\{1,\ldots,n\}$and (2b) a set of corresponding assigned values $l_i$. One can define a question $q_{v}$ that corresponds to the specific querying of a variable in the context of a dialog $p(l_i | q_{v_i}, d)$. In such context, a dialog state tracking task consists in determining for each variable v, $l^* = \argmax_{l_i\in L} p(l_i | q_{v_i}, d)$, with $L$ the specific domain of expression of a variable $v_i$. 

In addition to the actual dataset, we propose to investigate four general reasoning tasks using {\it DSTC-2} dataset as a starting point. In such way, we leverage the dataset of {\it DSTC-2} to create more complex reasoning task than the ones present in the original dialogs of the dataset by performing rule-based modification over the corpus. Obviously, the goal is to develop resolution algorithms that are not dedicated to a specific reasoning task but inference models that will be as generic as possible. In the rest of the section, each of the reasoning tasks associated with dialog state tracking are described and the generation protocol is explained with examples.

{\bf Factoid Questions} :  This first task corresponds to the current formulation of dialog state tracking. It consists of questions where a previously given a set of supporting facts, potentially amongst a set of other irrelevant facts, provides the answer. This kind of task was already employed in \cite{WestonCB14} in the context of a virtual world. In that sense, the result obtained to such task are comparable with the state of the art approaches. 

{\bf Yes/No Questions} :  This task tests the ability of a model to answer true/false type questions like {\it ``Is the food italian ?''}. The conversion of a dialog to such format is deterministic regarding the fact that the utterances and corresponding true states are known at each utterance of a given dialog. 

{\bf Indefinite Knowledge} : This task tests a more complex natural language construction. It tests if statements can be models in order to describe possibilities rather than certainties, as proposed in \cite{WestonCB14}. In our case, the answer will be ``maybe'' to the question {\it ``Is the price-range required moderate ?''} if the slot hasn't been mentioned yet throughout the current dialog. In the case of state tracking, it will allow to seamlessly deal with unknown information about the dialog state. Concretly, this set of questions and answers are generated has a super-set of the Yes-No Questions set. First, sub-dialog starting from the first utterance of a given dialog are extracted under the condition that a given slot is not informed in the corresponding annotation. Then, a question-answering question is generated.

{\bf Counting and Lists/Sets} :  This last task tests the capacity of the model to perform simple counting operations, by asking about the number of objects with a certain property, e.g. {\it ``How many area are requested ?''}. Similarly, the ability to produce a set of single word answers in the form of a list, e.g. {\it ``What are the area requested ?''} is investigated. Table \ref{tab:example} give an example of each of the question type presented below on a dialog sample of {\it DSTC-2} corpus. 

{\bf Inference procedure}: Concretely, the current set of utterances of a dialog will be placed into the memory using sentence based encoding and the question will be encoded as the controller state at $t=1$. The answer will be produced using a Softmax operation over the answer vocabulary that is supposed fixed. We consider this hypothesis valid in the case of factoid and list questions because the set of value for a given variable is often considered known. In the cases of Yes/No and Indefinite knowledge question, \{Yes, No, Maybe\} are added to the output vocabulary. Following \cite{WestonCB14}, a list-task answer will be considered as a single element in the answer set and the count question. A possible alternative would be to change the activation function used at the output of the \memnn from softmax activation function to a logistic one and to use a categorical cross entropy loss. A drawback of such alternative would be the necessity of cross-validating a decision threshold in order to select a eligible answers. Concerning the individual numbers for the count question set, the numbers founded on the training set are added into the vocabulary.  

\begin{table*}[ht!]
\centering
\resizebox{.72\textwidth}{!}{
\begin{tabular}{|l|l|l|}
\hline
{\bf Index} & {\bf Actor} & {\bf Utterance} \\
\hline\hline
1 &Cust & Im looking for a cheap restaurant in the west or east part of town. \\
2 &Agent& Thanh Binh is a nice restaurant in the west of town in the cheap price range. \\
3 &Cust &What is the address and post code. \\
4 &Agent &Thanh Binh is on magdalene street city centre.\\
5 &Cust &Thank you goodbye.  \\
\hline\hline
6 &  \multicolumn{2}{l|}{{\bf Factoid Question} What is the pricerange ? Answer: \{Cheap\} } \\
7 &  \multicolumn{2}{l|}{{\bf Yes/No Question} Is the Pricerange Expensive ? Answer: \{No\} } \\
8 &  \multicolumn{2}{l|}{{\bf Indefinite Knowledge} Is the FoodType chinese ? Answer: \{Maybe\} } \\
8 &  \multicolumn{2}{l|}{{\bf Listing task} What are the areas ? Answer: \{West,East\} } \\
\hline
\end{tabular}
}
\caption{{\bf } : Dialog state tracking question-answering examples from {\it DSTC2} dataset}
\label{tab:example}
\end{table*}

We believe more reasoning capabilities need to be explore in the future, like spacial and temporal reasoning or deduction as suggested in \cite{WestonBCM15}. However, it will probably need the development of a new dedicated ressource. Another alternative could be to develop a question-answering annotation task based on a dialog corpus where reasoning task are present. The closest work to our proposal that can be cited is \cite{BordesW16}. In this paper, the authors defines a so-called End-to-End learnable dialog system to infer an answer from a finite set of eligible answers w.r.t the current list of utterances of the dialog. The authors generate $5$ artificial tasks of dialog. However the reasoning capabilities are not explicitly addressed and the author explicitly claim that the resulting dialog system is not satisfactory yet. Indeed, we believe that having a proper dialog state tracker where a policy is built on top can guarantee dialog achievement by properly optimizing a reward function throughout a explicitly learnt dialog policy. In the case of proper end-to-end systems, the objective function is still not explicitly defined \cite{SerbanLCP15} and the resulting systems tend to be used in the context of chat-oriented and non-goal oriented dialog systems. In the next section, we present experimental details and results obtained on the basis of the {\it DSTC-2} dataset and its conversion to the four mentioned reasoning tasks. 
\section{Experiments}
\label{sec:xp}

\subsection{Dataset and Data Preprocessing}

In the {\it DSTC-2} dialog corpus, a user queries a database of local restaurants by interacting with a dialog system. A dialog proceeds as follows: first, the user specifies constraints concerning the restaurant. Then, the system offers the name of a restaurant that satisfies the constraints. Finally, the user accepts the offer and requests additional information about the accepted restaurant. In this context, the dialog state tracker should be able to track several types of information that compose the state like the geographic area, the food type and the price range slots. In order to make comparable experiments, subdialogs generated from the first utterance to each utterance of each dialog of the corpus have been generated. The corresponding question-answer pairs have been generated using the annotated state for each of the subdialog. In the case of factoid question, this setting allows for fair comparison at the utterance-level state tracking gains with the prior art. The same protocol has been adopted for the generated reasoning task. In that sense, the tracker task consists in finding the value $l^*$ as defined in Section \ref{sec:model}. In the overall dialog corpus, Area slot counts 5 possible values, Food slot counts 91 possible values and Pricerange slot counts 3 possible values. In order to exhibit reasoning capability of the proposed model in the context of dialog state tracking, three other dataset have been automatically generated from the dialog corpus in order to support $3$ capabilities of reasoning described in Section \ref{sec:model}. Dialog modification has been required for two reasoning tasks, {\it List} and {\it Count}. Two types of rules have been developed to automatically produce modified dialogs. On a first hand, string matching has been performed to determine the position of a slot values in a given utterance and an alternative statement has been produced as a substitution. For example, the utterance {\it ``I'm looking for a chinese restaurant in the north''} can be replaced by {\it ``I'm looking for a chinese restaurant in the north or the west of town''.} A second type of modification has been performed in an inter-utterance fashion. For example, assuming a given value ``north'' has been informed in the current state of a given dialog, one can add lately in the dialog a remark like ``I would also accept a place east side of town''. This kind of statement tends to not affect the overall flow of the dialog and allows to add richer semantic to the dialog. In the future, we plan to develop a richer set of generation procedures to augment the dataset. Nevertheless, we believe this simple dialog augmentation strategy allows to exhibit the competency of the proposed model beyond factoid questions.

\subsection{Training Details}
\label{sec:trainingdetails}

As suggested in \cite{SukhbaatarSWF15}, $10\%$ of the set was held-out to form a validation set for hyperparameter tuning. Concerning the utterance encoding, we use the so-called {\it Temporal Encoding} technique. In fact, reading tasks require some notion of temporal context. To enable the model to address them, the memory vector is modified as such $m_i = \sum_j Ax_{ij} + T_A(i)$, where $T_A(i)$ is the $i^{th}$ row of a dedicated matrix $T_A$ that encodes temporal information. The output embedding is augmented in the same way with a matrix $T_c$ (e.g. $c_i = \sum_j Cx_{ij} + T_C (i)$). Both $T_A$ and $T_C$ are learned during training in an end-to-end fashion. They are also subject to the same sharing constraints as $A$ and $C$. The embedding matrix $A$ and $B$ are initialized using GoogleNews word2vec embedding model \cite{MikolovSCCD13}. Also suggested on \cite{SukhbaatarSWF15}, utterances are indexed in reverse order, reflecting their relative distance from the question so that $x_1$ is the last sentence of the dialog. Furthermore, adjacent weight tying schema has been adopted. Learning rate $\eta$ is initially assigned a value of $0.005$ with exponential decay applied every $25$ epochs by $\eta/2$ until $100$ epochs are reached. Then, {\it linear start} is used in all our experiments as proposed by  \cite{SukhbaatarSWF15}. More precisely, the $\softmax$ function in each memory layer is removed and re-inserted after $20$ epochs. Batch size is set to $16$ and gradients with an $L_2$ norm larger than $40$ are divided by a scalar to have norm $40$. All weights are initialized randomly from a Gaussian distribution with zero mean and $\sigma = 0.1$. In all our experiments, we have tested a set of the embedding size $d \in \{20,40,60\}$. After validation, each model uses a $5$-hops depth configuration.

\subsection{Experimental results}

\begin{table*}[ht!]
\small
\centering
\resizebox{\textwidth}{!}{
\begin{tabular}{|l|l|l|l|l|l|l|}
\hline
{\bf Locutor } & {\bf Utterance }& {\bf Hop 1 } & {\bf Hop 2 } & {\bf Hop 3 } & {\bf Hop 4 } & {\bf Hop 5 } \\
\hline
Cust& Im looking for a cheap restaurant that serves chinese food	&0.00	&\cellcolor{blue!18}0.18	&0.11	&0.04&	0.00 \\
\hline
Agent& What part of town do you have in mind	&\cellcolor{blue!33}0.33	&\cellcolor{blue!30}0.30&	0.00	&0.00&	0.00 \\
\hline
Cust& I dont care	&0.00	&0.00&	\cellcolor{blue!17}0.17	&\cellcolor{blue!37}0.37&	\cellcolor{blue!100}1.00 \\
\hline
Agent& Rice house serves chinese food in the cheap price range&	\cellcolor{blue!1}0.01	&0.00	&0.00	&0.00&	0.00 \\
\hline
Cust& What is the address and telephone number	&\cellcolor{blue!58}0.58	&\cellcolor{blue!9}0.09&	\cellcolor{blue!1}0.01	&0.00	&0.00 \\
\hline
Agent& Sure rice house is on mill road city centre	&\cellcolor{blue!3}0.03	&0.00	&0.00&	0.00&	0.00 \\
\hline
Cust& Phone number	&0.00&0.00&	0.00&	0.00&	0.00 \\
\hline
Agent& The phone number of rice house is 765-239-09	&\cellcolor{blue!2}0.02&	\cellcolor{blue!1}0.01&	0.00&	0.00	&0.00 \\
\hline
Cust& Thank you good bye	&\cellcolor{blue!2}0.02&	\cellcolor{blue!42}0.42	&\cellcolor{blue!71}0.71	&\cellcolor{blue!59}0.59&	0.00 \\
\hline
\multicolumn{7}{|l|}{{\bf What is the area ? Answer: dontcare}} \\
\hline
\end{tabular}
}
\vspace{0.3cm}
\caption{Attention shifting example for the {\it Area} slot from {\it DSTC2} dataset, the values corresponds the $p_i$ values affected to each memory block $m_i$ at each hop of the \memnn}
\label{tab:mem3}
\end{table*}

Table \ref{tab:results} presents tracking accuracy obtained for three variables of the {\it DSTC2} dataset formulated as {\it Factoid Question} task. We compare with two established utterance-level discriminative neural trackers, a Recurrent Neural Network (RNN) model \cite{Henderson2014d} and the Neural Belief Tracker \cite{SWTY16}. As suggested in this last work, the first RNN baseline model uses no semantic (i.e. synonym) dictionary, while the improved baseline uses a hand-crafted semantic dictionary designed for the {\it DSTC2} ontology. In this context, a \memnn model allows to obtain competitive results with the most close, non-memory enhanced, state of the art approach of recurrent neural network with word embedding as prior knowledge.

\begin{table}[ht!]
\centering
\resizebox{0.45\textwidth}{!}{
\begin{tabular}{|l|l|l|l|l|}
\hline
Model & Area & Food & Price & Joint \\
\hline
RNN - no dict. & 0.92 & 0.86 & 0.86 & 0.69 \\
RNN + sem. dict.& 0.91 & 0.86 & 0.93 & 0.73  \\
NBT-DNN & {\bf0.90} & 0.84 & 0.94  & 0.72  \\
NBT-CNN & {\bf0.90} & 0.83 & 0.93 & 0.72  \\
\hline 
MemN2N($d=40$) & 0.89 & {\bf 0.88} & {\bf 0.95} & {\bf  0.74} \\
\hline
\end{tabular}
}
\vspace{0.3cm}
\caption{{\bf One supporting fact task} : Acc. obtained on DSTC2 test set}
\label{tab:results}
\end{table}

As a second result, Table \ref{tab:resultsreason} presents the performance obtained for the four reasoning tasks. The obtained results lead us to think that \memnn are a competitive alternative for the task dialog state tracking but also increase the spectrum of definition of the general dialog state tracking task to machine reading and reasoning. In the future, we believe new reasoning capabilities like spacial and temporal reasoning and deduction should be exploited on the basis of a specifically designed dataset.

\begin{table}[tb!]
\centering
\resizebox{0.45\textwidth}{!}{
\begin{tabular}{|l|l|l|l|l|l|}
\hline
{\bf Variable} & {\bf d} & {\bf Yes-No} & {\bf I.K.} & {\bf Count.} & {\bf List.}  \\
\hline
 & 20 & {\bf 0.85} & 0.79 & 0.89 & 0.41\\ 
Food & 40 &   0.83 & {\bf 0.84} &  0.88 &  {\bf 0.42}\\ 
 & 60 & 0.82 & 0.82 &  {\bf 0.90} & 0.39\\ 
\hline
 & 20 &  0.86  & 0.83 &  0.94 & {\bf 0.79} \\ 
Area & 40 &  {\bf 0.90}  & 0.89 & {\bf 0.96}  & 0.75 \\ 
 & 60 & 0.88   & {\bf 0.90} &   0.95 & 0.78 \\ 
\hline
& 20 & {\bf  0.93}  & {\bf 0.86} & {\bf 0.93} & {\bf 0.83} \\ 
PriceRange & 40 &  0.92  & 0.85 &  0.90 & 0.80 \\ 
& 60 & 0.91 & 0.85 &  0.91 & 0.81 \\ 
\hline
\end{tabular}
}
\caption{{\bf Reasoning tasks} : Acc. on DSTC2 reasoning datasets}
\label{tab:resultsreason}
\end{table}
\section{Conclusion and Further Work}

This paper describes a novel method of dialog state tracking based on the paradigm of machine reading and solved using \memnn, a memory-enhanced neural network architecture. In this context, a dataset format inspired from the current datasets of machine reading tasks has been developed for this task. It is the first attempt to solve this classic sub-problem of dialog management in such way. Beyond the experimental results presented in the experimental section, the proposed approach offers several advantages compared to state of the art methods of tracking. First, the proposed method allows to perform tracking on the basis of segment-dialog-level annotation instead of utterance-level one that is commonly admitted in academic datasets but tedious to produce in a large scale industrial environment. Second, we propose to develop dialog corpus requiering reasoning capabilities to exibit the potential of the proposed model. In future work, we plan to address more complex tasks like spatial and temporal reasoning, sorting or deduction and experiment with other memory enhanced inference models. Indeed, we plan to experiment and compare the same approach with Stacked-Augmented Recurrent Neural Network \cite{JoulinM15} and Neural Turing Machine \cite{GravesWD14} that sounds also promising for these family of reasoning tasks. 

\bibliographystyle{eacl2017}
\bibliography{biblio}

\end{document}